\documentclass[letterpaper, 10 pt, conference]{ieeeconf}  

\IEEEoverridecommandlockouts                              

\pdfoutput=1\relax                   
\usepackage{graphicx}                
\DeclareGraphicsExtensions{.pdf,.png,.jpg,.jpeg} 

\usepackage{microtype}
\usepackage{mathptmx}
\usepackage{amsfonts}
\usepackage{amsmath,amssymb}
\usepackage{bm}
\usepackage{subfigure}
\usepackage{caption}
\usepackage{tabu}                      
\usepackage{booktabs}                  
\usepackage{multirow}
\usepackage{graphics} 
\graphicspath{{figures/}{pictures/}{images/}{./}} 

\title{\LARGE \bf
VIP-SLAM: An Efficient Tightly-Coupled RGB-D Visual Inertial Planar SLAM
\vspace{-0.25cm}
}
 
 \author{Danpeng Chen$^{1,2,3}$\quad Shuai Wang$^{2,3}$\quad Weijian Xie$^{1,2}$\quad Shangjin Zhai$^{2}$, \\Nan Wang$^{2,3}$\quad Hujun Bao$^{1}$ \quad Guofeng Zhang$^{1*}$
 \\$^{1}$ State Key Lab of CAD\&CG, Zhejiang University \quad  $^{2}$SenseTime Research \quad  $^{3}$Tetras.AI
 \thanks{*Corresponding author: Guofeng Zhang (zhangguofeng@zju.edu.cn)}%
 \thanks{This work was partially supported by the National Key Research and Development Program of China under Grant 2020AAA0105900. All the authors are also affiliated with ZJU-SenseTime Joint Lab of 3D Vision.}
 \vspace{-0.5cm}
 }
\begin{document}
\maketitle

\begin{abstract}
In this paper, we propose a tightly-coupled SLAM system fused with RGB, Depth, IMU and structured plane information. Traditional sparse points based SLAM systems always maintain a mass of map points to model the environment. Huge number of map points bring us a high computational complexity, making it difficult to be deployed on mobile devices. On the other hand, planes are common structures in man-made environment especially in indoor environments. We usually can use a small number of planes to represent a large scene. So the main purpose of this article is to decrease the high complexity of sparse points based SLAM. We build a lightweight back-end map which consists of a few planes and map points to achieve efficient bundle adjustment (BA) with an equal or better accuracy. We use homography constraints to eliminate the parameters of numerous plane points in the optimization and reduce the complexity of BA. We separate the parameters and measurements in homography and point-to-plane constraints and compress the measurements part to further effectively improve the speed of BA. We also integrate the plane information into the whole system to realize robust planar feature extraction, data association, and global consistent planar reconstruction. Finally, we perform an ablation study and compare our method with similar methods in simulation and real environment data. Our system achieves obvious advantages in accuracy and efficiency. Even if the plane parameters are involved in the optimization, we effectively simplify the back-end map by using planar structures. The global bundle adjustment is nearly 2 times faster than the sparse points based SLAM algorithm.
\end{abstract}

\begin{figure*}[]
    \setlength{\abovecaptionskip}{-0.0cm}
    \setlength{\belowcaptionskip}{-0.7cm}
    \hsize=\textwidth
    \centering
    \includegraphics[width=0.75\linewidth]{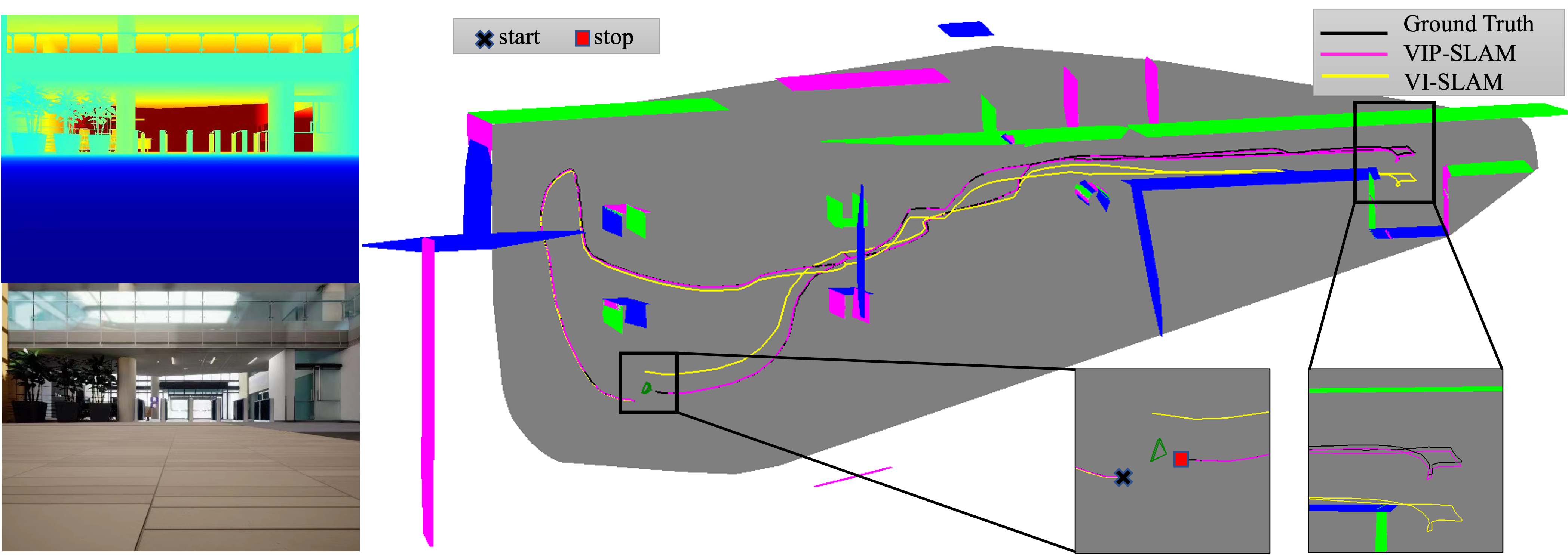}
    \captionsetup{font={footnotesize}}
    \caption{Trajectories and reconstructed planes from sequence 02. Top left is depth image, bottom left is RGB image. Right image: The gray area is horizontal plane. Pink, green and blue areas are different vertical planes. Black, pink and yellow trajectories are Ground Truth, our VIP-SLAM and normal VI-SLAM.}
    \label{fig:plane_pose}
\end{figure*}
\thispagestyle{empty}
\pagestyle{empty}

\section{INTRODUCTION}
6DoF motion estimation in unknown environments is a very critical and challenging technology for many applications, such as robot navigation, autonomous driving and AR/VR. 
To solve the indoor positioning problem, in recent years, many researchers have concentrated on using onboard sensors for real-time Simultaneous Localization and Mapping (SLAM).

At present, many visual SLAM~\cite{mur2015orb,mur2017orb,liu2017robust,newcombe2011kinectfusion,whelan2016elasticfusion,whelan2012kintinuous} can achieve high positioning accuracy. However, only relying on external scene information such as vision and geometry is easily affected by the external environment. 
To solve the problem of environment dependence, a lot of visual inertial odometry~\cite{wu2015square,forster2015imu,qin2018vins} and visual inertial SLAM~\cite{shan2019rgbd,campos2021orb,mur2017visual} are proposed. They integrate IMU information that does not depend on the external environment into the system. 

Although the existing visual inertial systems have achieved very high accuracy and robustness, there are still many problems in practical applications. The visual inertial odometry has cumulative drift. Visual inertial SLAM needs to optimize many sparse map points, keyframe poses and IMU states, and these high-complexity problems are difficult to solve in real-time on mobile devices. 
There are many planar structures in the indoor environment.
The plane constraints have two major advantages. 1) The plane usually is a larger structure feature, and these large features may effectively suppress the cumulative drift of SLAM; 2) Compared with feature points, plane can use fewer parameters to model a larger environment. Therefore, how to use these higher-level plane features to improve the localization accuracy and stability of the SLAM and reduce the system complexity are significant problems that needs to be solved.

So far, there are many literatures~\cite{hsiao2018dense,ma2016cpa,taguchi2013point,zhang2019point,rosinol2019kimera,arndt2020points,li2020structure,yunus2021manhattanslam,joo2020linear,li2020co} using line and planar features to improve the accuracy and robustness of SLAM in challenging environments. But most of them are just odometry or pure visual SLAM, and there is no literature to propose a complete RGB-D inertial plane tightly-coupled SLAM. 
There is also work~\cite{li2020co} discussing the use of planes to reduce the dimension of the Hession matrix and improve the solving speed of BA, but it is the only odometry, and re-parameterization does not reduce the number of constraints.

In summary, many previous works have shown that IMU helps to improve the robustness of the system, and structured plane helps to improve the accuracy and robustness of the system. Moreover, compared with line and point features, the plane can use fewer parameters to model the environment. Based on this, we make full use of the characteristics of multiple sensors to propose a highly robust and precise system, which integrates IMU, RGB, Depth, and Plane information.
There are three contributions in this article:
\begin{itemize}
    \item We are the first to propose a complete tightly-coupled multi-sensor fusion SLAM system to fuse RGB, Depth, IMU, and structured plane information. 
    All the information are integrated into an unified non-linear optimization framework, which jointly optimizes the parameters of keyframe poses, IMU states, points, and planes.
    \item We introduce plane information to reduce the number of map points and speed up the optimization of BA. We use homography to remove the states of the point in the optimization and compress multiple constraints into one at the same time. These measures reduce the optimization time. The Fig \ref{fig:homography0} and \ref{fig:homography1} show the process.
    \item The plane information is integrated into the entire SLAM system to realize high-precision tracking. 
    We use pure geometric single-frame point-to-plane constraints to improve the accuracy and stability of plane estimation in textureless scenes. Moreover, we convert the reprojection of the plane point to homography constraints to establish the relationship between multiple frames and planes to further correct the drift.
    Fig.~\ref{fig:plane_pose} shows comparison of trajectories and reconstructed planes.
\end{itemize}
We evaluate our system both on simulation and real data. The data includes slow and fast movements, low texture, small indoor rooms and large scenes. Experimental results show that our system not only achieves higher accuracy and robustness but also has a significant improvement in the efficiency of solving BA.

\section{RELATED WORK}
\textbf{RGB-D Inertial SLAM}: 
In the past few years, RGB-D sensors have become more and more popular. 
Many methods~\cite{newcombe2011kinectfusion,whelan2012kintinuous,whelan2016elasticfusion,liu2017robust,mur2017orb} are proposed to use RGB-D cameras for real-time 3D dense reconstruction or SLAM. However, traditional RGB-D SLAM easily suffer from the robustness problem in special scenarios, such as fast motion, weak textures, dynamic environments, flat white walls, etc. To reduce the dependence on vision and geometry, researchers proposed a few RGB-D inertial SLAM systems. 
\cite{laidlow2017dense} proposes to integrate IMU and RGB-D information to achieve the first tightly-coupled RGB-D inertial SLAM system.
Their experiments showed the advantages of their proposed system in fast motion, weak texture, and weak geometric scenes.
VINS-RGBD~\cite{shan2019rgbd} integrates depth information to the monocular visual inertial system VINS-Mono~\cite{qin2018vins}. They use depth information to speed up system initialization, stabilize system scale and improve system accuracy and robustness. DPI-SLAM~\cite{hsiao2018dense} proposes the first plane inertial SLAM system which tightly couples IMU, visual odometry (VO), and plane information. However, the loose coupling of IMU and VO may encounter the same robustness issues as RGB-D SLAM.
In ORB-SLAM3~\cite{campos2021orb}, inertial constraints and the methods of multi-maps are fused into ORB-SLAM2~\cite{mur2017orb}. However, ORB-SLAM3 only supports monocular/stereo visual inertial SLAM or RGB-D SLAM.

\textbf{Plane Based SLAM}: 
Some researchers have explored ways to improve the accuracy of SLAM systems using planar structures. \cite{ma2016cpa} proposes a RANSAC-based registration method for localization based on points and planes. \cite{taguchi2013point} uses the global plane model to reduce the RGB-D SLAM drift. Kimera~\cite{rosinol2019kimera} proposes to use 2D image Delaunay triangulation and corresponding sparse point cloud to reconstruct 3D Mesh. 
SP-SLAM~\cite{zhang2019point} adds features to ORB-SLAM2, where parallel and perpendicular plane constraints also are included. \cite{arndt2020points} forces the points on the plane to fall exactly on the plane, and add new feature points and plane constraints based on this assumption.
KDP-SLAM~\cite{hsiao2017keyframe} develops a fast dense planar SLAM system, which optimizes the keyframe poses and planes in a global factor graph. But, the loose coupling of planes and VO may lead to a decrease in the accuracy of motion estimation in scenes with one or two planes.
Some works model the environment as the Manhattan or Atlanta world and use this assumption to reduce the cumulative localization error in rotation. \cite{li2020structure} decouples rotation and translation estimation based on Manhattan assumption, and combines point, line, and surface features to improve the accuracy of translation. 
\cite{yunus2021manhattanslam}'s system is also based on the Manhattan world hypothesis, but it can support seamless switching to the non-Manhattan world. 
These world hypothesis algorithm can work well under special scenes. However, it does not work well under complex environments.
There are also some works focused on improving the speed of plane bundle adjustment (PBA). \cite{zhou2020efficient,zhou2021lidar,zhoupi} propose to merge multiple measurements related to the same state into one constraint, which significantly improve the efficiency of PBA. Inspired by these works, we also merge point-to-plane constraints and extend them to homography observations. \cite{li2020co} forces the points associated with the plane to fall on the plane and uses the plane and the corresponding anchor pose to represent the point. This form of representation eliminates the point state and reduces the dimensionality of the Hessian matrix, which will increase BA solved speed.

\section{OVERVIEW}
\subsection{System Overview} 
\begin{figure}[htb]
    \setlength{\belowcaptionskip}{-0.5cm}
    \centering
    \includegraphics[width=0.95\linewidth]{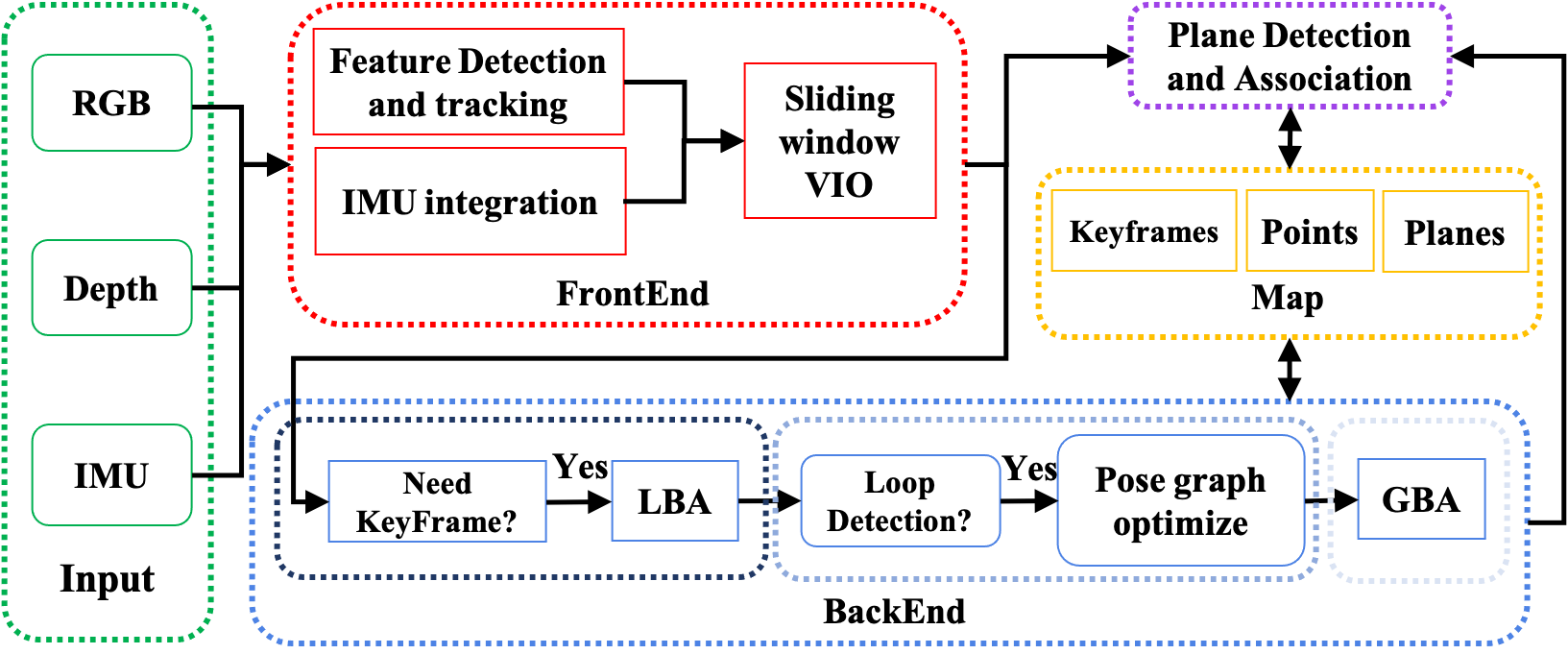}
    \captionsetup{font={footnotesize}}
    \caption{The pipeline of the proposed system.}
    \label{fig:system}
\end{figure}

The overview of the proposed system is shown in Fig. \ref{fig:system}.
Our system takes RGB, Depth, and IMU as input and has three main components including front end, plane module, and back end.

The front-end module is a VIO system based on a sliding window and estimates 6-DoF poses in real-time. Our VIO is similar to~\cite{wu2015square}, except that we do not consider the SLAM features and add additional depth measurements.

The plane module receives the front-end and back-end data as input. The high-frequency front-end information is only used to expand the plane. The low-frequency but high-precision back-end information is accepted for new plane detection, plane expansion, point-to-plane association, and plane-to-plane merging. Since most planes of indoor environment are horizontal or vertical, we only consider the detection of horizontal and vertical planes. However, the fusion optimization about planes is suitable for general planes.

The back-end module uses local bundle adjustment (LBA) or global bundle adjustment (GBA) to jointly optimize planes, points, IMU states and keyframe poses and 
corrects the drift of the front-end pose.

\subsection{Notation}
We first define notations that used throughout the paper.
We consider $(\cdot)^W$ as the world frame. 
$(\cdot)^C$ is the camera frame and $(\cdot)^I$ is the IMU frame. 
We represent a pose by $\bm{T}\in SE(3)$ which consists of rotation $\bm{R}\in SO(3)$ and translation $\bm{t}\in R^3$. 
We use $\bm{\pi}=[\bm{n};d]^T$ to represent a plane, where $\bm{n}$ is the plane normal, and $d$ is the distance between the origin and plane. We adopt CP~\cite{geneva2018lips} vector to parameterize plane $\bm{\pi}$, $i.e., \bm{\eta} =\bm{n}d$. 
$\bm{X}$ are the states need to be estimated, including poses, velocities, IMU bias, points and plane landmarks.

\section{Front End}
\textbf{Feature Detection and Matching} 
When a new image is received, we detect ORB feature~\cite{mur2015orb} points and compute corresponding descriptors. First, we use KLT to track them from the last image to the current image. Then, we project features that have 3D information onto the current image and use the Hamming distance to find the closest ORB feature point. The remaining features are matched by using the result of KLT tracking as the initial value to find the optimal observation. Finally, we use the RANSAC-based fundamental matrix to remove outliers.

\textbf{Motion Estimation}
Motion estimation is a sliding window based VIO, which tightly integrates RGB, depth, and IMU. Our VIO is similar to~\cite{wu2015square}, which uses a square root inverse filter to fuse all measurements. The major difference is that we do not consider the SLAM features and add depth information to visual measurements. We will descript visual measurements with depth in detail in Section \ref{point_measurement}.

\section{Plane Detection and Association}\label{plane_module}
\subsection{Plane Detection and Merging}
\begin{figure}[htb]
    \setlength{\belowcaptionskip}{-0.7cm}
    \centering
    \includegraphics[width=0.8\linewidth]{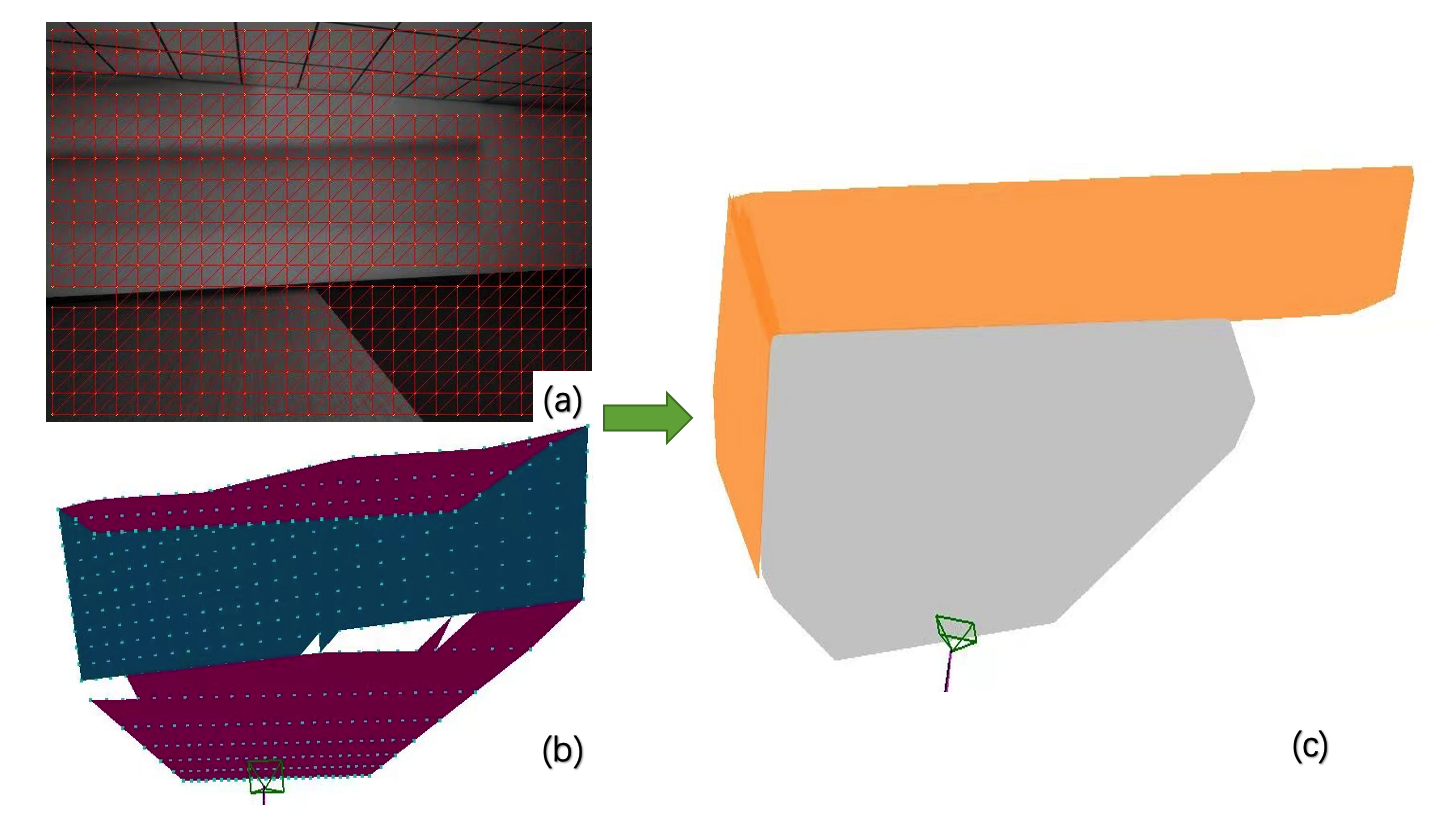}
    \captionsetup{font={footnotesize}}
    \caption{Plane detection: (a) 2D mesh. (b) 3D mesh, where red and blue areas indicate horizontal and vertical meshes respectively. (c) Planes where vertical planes are yellow, and horizontal plane is grey.}
    \label{fig:plane-detect}
\end{figure}

\textbf{Plane Detection}
The plane module only detects planes with back-end data. Once a plane is detected, the plane module expands the plane with front-end and back-end data and associates the plane with map landmarks. 
Like~\cite{rosinol2019kimera}, we use Delaunay triangulation to create 3D mesh and histogram to detect planes. We only detect vertical and horizontal planes, by checking out whether the mesh normal is vertical or parallel to the gravity. 
Our plane module employs some additional methods to improve the plane accuracy. When a plane is detected from histogram, we will refine its parameters with the data and 3D plane points in the histogram, instead of using the scale value of the histogram directly. For horizontal plane, we just set 
$\bm{n} = [0, 0, 1]^T$ and the plane distance is the average of the z axis of plane points. For vertical plane, we set $\bm{n} = [n_x, n_y, 0]^T$, and the vertical plane parameters can be refined with the following Equation:
\begin{equation}
    \begin{split}
        \begin{aligned}
          \left [ \begin{matrix} \bar{\bm{P}}^W_{f_1}, 
                                 \bar{\bm{P}}^W_{f_2}, 
                                 \cdot \cdot \cdot, 
                                 \bar{\bm{P}}^W_{f_n} \end{matrix} \right ]^T
                              \cdot 
            \left [\begin{matrix}
               n_x/d \\
               n_y/d \\
            \end{matrix} \right ] = \left [ \begin{matrix} -1, 
                              -1, 
                              \bm{\cdot} \bm{\cdot} \bm{\cdot}, 
                              -1 \end{matrix} \right ]^T
        \end{aligned}
    \end{split}
    \label{detect_plane}
\end{equation}
where $n$ is the number of plane points, $ \bm P^W_{f_k} $ is the position of $k^{th}$ landmark under the world frame, and $ \bar{\bm{P}}^W_{f_k} = {\left [\bm{P}^W_{f_k}(0), \bm{P}^W_{f_k}(1) \right ]^T} $. We use QR decomposition to solve the Equation (\ref{detect_plane}).
When plane parameters are detected, the plane can associate 3D mesh by angle and distance. Our angular and distance thresholds are about 10 degrees and 5 centimeters between the plane and 3D mesh. Fig.~\ref{fig:plane-detect} shows this process.

\textbf{Plane Merging}
The plane merging has 2 strategies. First, we check whether a plane satisfies a certain angle and distance threshold with others. Once we find satisfied planes, secondly, we will check the boundary points of satisfied planes whether contain each other. Our angular and distance thresholds are 10 degrees and 10 centimeters. Plane merging occurs when a new plane is detected or old planes are adjusted. 

\subsection{Point and Plane Association}
We use 3D meshes 
to associate more map points with planes. Once a 3D mesh from the depth map has been associated with a plane, we will find its 2D mesh. If both the 2D coordinate of a map point is in the 2D mesh and the distance from the map point to the plane is less than 10 centimeters, the map point will be added into the candidate set associated with the plane. If a point on the candidate set is observed in more than 3 keyframes, we will check its geometric consistency. We calculate the reprojection error of the point, force the point to be associated with the plane, and then calculate another reprojection error. If the two reprojection errors are similar, and the maximum reprojection error is lower than a certain threshold, we consider the point to be a plane point. If a point fails to pass geometric consistency many times, it will be removed from the candidate point set.

\begin{figure}[htb]
    \centering
    \captionsetup{font={footnotesize}}
    \vspace{-0.5cm}  
    \setlength{\belowcaptionskip}{-0.8cm}
    \subfigure[Homography Factor]{
    \centering
    \setlength{\belowcaptionskip}{-0.5cm}
    \includegraphics[width=0.35\linewidth]{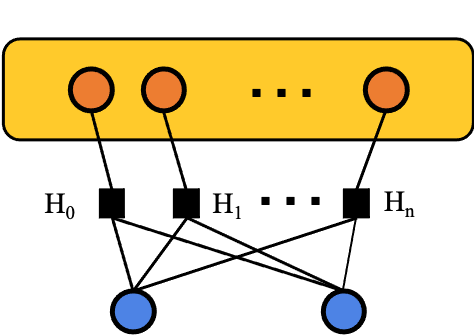}
    \label{fig:homography0}
    }%
    \subfigure[Compressed Homography Factor]{
    \centering
    \setlength{\belowcaptionskip}{-0.5cm}
    \includegraphics[width=0.35\linewidth]{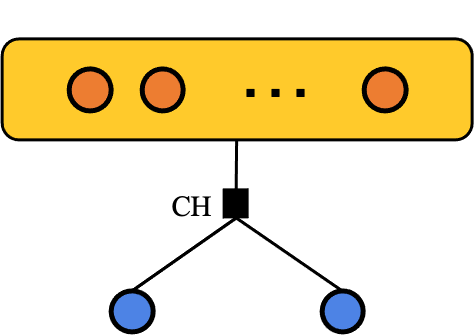}
    \label{fig:homography1}
    }%
    
    \subfigure[Plane Point Optimization Problem]{
    \centering
    \includegraphics[width=0.9\linewidth]{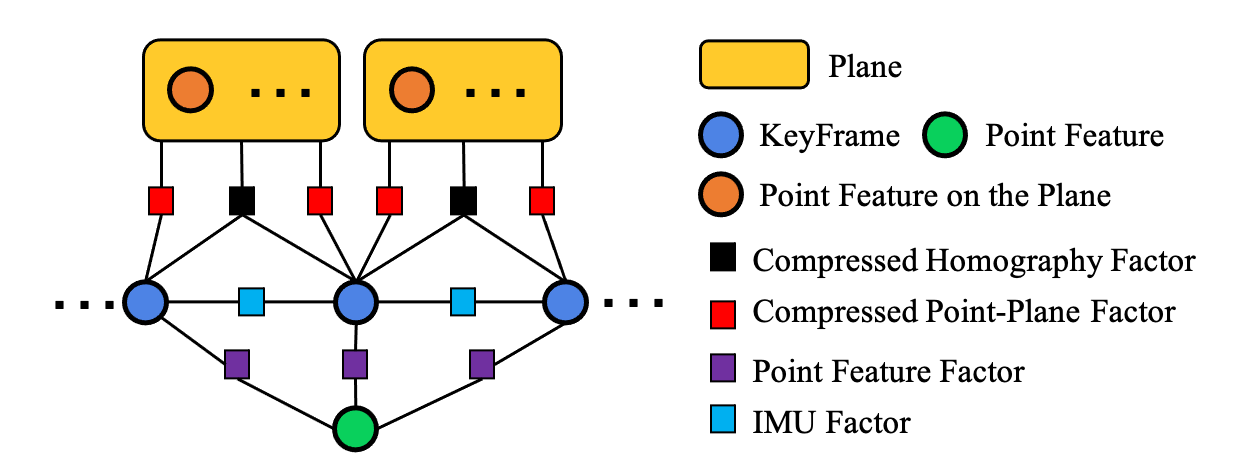}
    }%
    \centering
    \captionsetup{font={footnotesize}}
    \caption{Our optimization problem. 
    (a) and (b) are the factor graphs before and after homography constraints are compressed.
     (c) is our global bundle adjustment problem.}
    \label{fig:problem}
\end{figure}

\section{BACK END}
\subsection{Measurement}
\subsubsection{IMU and Point Feature Measurement}\label{point_measurement}
The IMU data between two continuous keyframes are processed by the preintegration method~\cite{forster2015imu,qin2018vins}. We define the cost term of preintegration based IMU data, which is the same as~\cite{qin2018vins}.

Since depth image is valid, we integrate depth information into visual point feature measurement. The projection and depth residuals for the $l^{th}$ feature observed in the $i^{th}$ keyframe is defined as:
{
\begin{equation}
    \begin{split}
        \begin{aligned}
            \bm{r}^C(\bm{X}) &= Proj(\bm{P}_l^{C_i}) - \bm{z}_{l_i} \\
            \bm{r}^{C_{\lambda}}(\bm{X}) &= (\bm{P}_l^{C_i}).z() - {\lambda}
        \end{aligned}
    \end{split}
\end{equation}}
where $\bm{P}_l^{C_i} = {\bm{R}_C^I}^T(\bm{R}_W^{I_i}(\bm{P}_l^W - \bm{t}_{I_i}^W) - \bm{t}_C^I)$ is the 3D position of $l^{th}$ feature in the $i^{th}$ keyframe. $\bm{R}^I_C$, $\bm{t}^I_C$ are rotation and translation from the IMU frame to the Camera frame. $Proj$ is the project function that project the point from camera coordinate to image coordinate.
$\bm{z}_{l_i}$ is the $l^{th}$ feature observation in the $i^{th}$ image. $.z()$ is the third component of the vector. 
${\lambda}$ is the corresponding depth obtained from the depth image.

\subsubsection{Compressed Homography Measurement}\label{compressed_h}
When the 3D map point is associated with the plane, we enforce the point must fall on the plane. So instead of using the common point-to-plane distance constraint, we use the homography matrix to constrain two keyframes and a plane. 
If the point on the plane $\bm{\pi}^W$ is observed by frames $i^{th}$ and $j^{th}$, we can write the following Equations of point-to-plane and reprojection:
\begin{equation}\label{homography_e0}
    \begin{split}
        \begin{aligned}
            {\bm{n}_{\pi}^W}^T(\bm{R}_{C_i}^W \bm p_i  {\lambda} + \bm{t}_{C_i}^W)+d_{\pi}^W &= 0 \\
            {\bm{R}_{C_j}^W}^T(\bm{R}_{C_i}^W \bm p_i {\lambda} + \bm{t}_{C_i}^W - \bm{t}_{C_j}^W) &=  \bm p_j s 
        \end{aligned}
    \end{split}
\end{equation}
We define $\bm p_i=(x_i,y_i,1)^T$ and $\bm p_j=(x_j,y_j,1)^T$ , where $(x_i,y_i,1)^T = \bm K^{-1}(u_i, v_i, 1)^T$ is the point feature observation on the normal plane of the frame $i^{th}$. $\bm K$ is the intrinsic matrix, and $(u_i, v_i)$ is a 2D image feature. ${\lambda}$ is the corresponding depth. $s$ is an unknown scale.
Combining the above Equations (\ref{homography_e0}), we can get:
{
\setlength\abovedisplayskip{0.5pt}
\setlength\belowdisplayskip{0.5pt}
\begin{equation}\label{homography_euqtion}
    \begin{split}
        \begin{aligned}
          \bm p_j s &= {\bm{R}_{C_j}^W}^T[\bm{I}-\frac{(\bm{t}^W_{C_i}-\bm{t}^W_{C_j}){\bm{n}_{\pi}^W}^T}{d_{\pi}^W+{\bm{n}_{\pi}^W}^T \bm{t}_{C_i}^W}]\bm{R}_{C_i}^W \bm p_i
 = \bm{H} \bm p_i 
        \end{aligned}
    \end{split}
\end{equation}
}
where $\bm{I}$ is an identity matrix, and $\bm{H}$ is a homography matrix. So the homography constraint and the reprojection constraint are equal when the point is on the plane.
\textbf{It is worth noting that homography does not require the 3D positions of point features}. In the BA problem, we convert the reprojection constraint to homography constraint, which is equivalent to remove many state variables of points on the plane. Finally, the efficiency of bundle adjustment will be greatly improved through the smaller and sparser Hessian matrix. There is also a similar work~\cite{li2020co} to remove the states of the plane points in optimization, but they use the reprojection representation method, it is difficult to compress multi-observation constraints, which limits the further improvement of the optimization speed. Homography associates the states of two keyframes and a plane. These three states may have many common observations. We merge these observations into one observation to further improve the optimization speed. 
Assume there are $N$ point features on the plane $\bm{\pi}$ observed by $i^{th}$ and $j^{th}$ keyframes.
According to Equation~(\ref{homography_euqtion}), the homography constraint Equation of $l^{th}$ point feature is:
\begin{equation} \label{homography_e1}
    \begin{split}
        \begin{aligned}
            s\left [ \begin{matrix}
 x_j\\
 y_j\\
1
\end{matrix} \right ] =\left [\begin{matrix}
  H_{11}&  H_{12}& H_{13}\\
  H_{21}&  H_{22}& H_{23}\\
  H_{31}&  H_{32}& H_{33}
\end{matrix}  \right ] \left [ \begin{matrix}
 x_i\\
 y_i\\
1
\end{matrix} \right ]
        \end{aligned}
    \end{split}
\end{equation}
We can eliminate unknown $s$ from the above Equation (\ref{homography_e1}) and get the residual function as:
\begin{small}
    \begin{equation}
        \begin{split}
            \begin{aligned}
                \bm{r}_{l}(\bm{X}) = &\left [ \begin{matrix}
      x_i&  y_i&  1&  0&  0&  0&  -x_ix_j&  -y_ix_j& -x_j\\
      0&  0&  0&  x_j&  y_j&  1&  -x_iy_j&  -y_iy_j& -y_j
    \end{matrix} \right ] \\
    &\left [ \begin{matrix}
    H_{11}&
    H_{12}&
    H_{13}&
    H_{21}&
    H_{22}&
    H_{23}&
    H_{31}&
    H_{32}&
    H_{33}
    \end{matrix} \right ]^T \\
      &= \bm{C}_l\bar{\bm{H}} 
            \end{aligned}
        \end{split}
    \end{equation}
\end{small}
The homography cost function of $N$ point features on the plane $\bm{\pi}$ is:
{\setlength\abovedisplayskip{0.5pt}
\setlength\belowdisplayskip{2.0pt}
\begin{small}
\begin{equation}
    \begin{split}
        \begin{aligned}
            \bm{C}^{ch}(\bm{X}) &= \frac{1}{2} \sum_{l=1}^{N} {\bm{r}_l(\bm{X})}^T \bm{r}_l(\bm{X}) 
                   = \frac{1}{2} {\bar{\bm{H}}}^T (\sum_{l=1}^{N} {\bm{C}_l}^T \bm{C}_l) {\bar{\bm{H}}} \\
                   &= \frac{1}{2} {\bar{\bm{H}}}^T \bm{G_h} {\bar{\bm{H}}} 
                   = \frac{1}{2} {\bar{\bm{H}}}^T \bm{L_h}\bm{L_h}^T {\bar{\bm{H}}}
        \end{aligned}
    \end{split}
\end{equation}
\end{small}}
where $\bm{G_h}=\bm{L_h}\bm{L_h}^T$ is the matrix decomposition. To ensure the stability of the solution, we use eigenvalue decomposition. $\bm{G_h}$ is a constant $9\times 9$ matrix during the optimization and only depends on visual observations. 
We can calculate the matrix in advance. Here, we have merged observations of $N$ point features on the plane $\bm{\pi}$ into one observation matrix $\bm{G_h}$. 
Fig.~\ref{fig:homography0} and \ref{fig:homography1} show this process.
Combining many cost functions into one cost function helps to improve the efficiency of bundle adjustment.
The jacobian of compressed homography cost function can be define as $\bm{J}^{ch}=\bm{L_h}^T \frac{\partial \bar{\bm{H}} }{\partial \bm{X}}$, and residue is $\bm{r}^{ch}=\bm{L_h}^T\bar{\bm{H}}$ 

\subsubsection{Compressed Point-to-Plane Measurement}\label{compressed_pp}

The homography measurement relies on feature matching of map point, which is easily affected by the lighting and texture of environments. Therefore, we add geometric constraint of single frame point cloud and plane association to enhance the accuracy of plane estimation and the stability of motion estimation under textureless scenes. Similar to the work related to the Lidar plane SLAM~\cite{zhoupi,zhou2021lidar,zhou2020efficient}, we use the compressed point-to-plane cost function. $\mathbb{P}_{li}$ is the set of $N$ points on the $l^{th}$ plane $\bm{\pi}_l^W$ observed in the $i^{th}$ keyframe. $\bm{p}_{ilk}$ is the $k^{th}$  3D point of $\mathbb{P}_{li}$ in world . The point-to-plane residue of $k^{th}$ plane point can be defined as:
{\setlength\abovedisplayskip{0.5pt}
\setlength\belowdisplayskip{0.5pt}
\begin{equation}
    \begin{split}
        \begin{aligned}
            \bm{r}_{ilk}(\bm{X})
                       = {\bm{\pi}_l^W}^T \bm{T}_i^W \left [ \begin{matrix}
                                                     \bm{p}_{ilk}\\
                                                     1
                                                    \end{matrix} \right ] 
        \end{aligned}
    \end{split}
\end{equation}
}
where $\bm{T}_i^W$ is the $i^{th}$ keyframe pose. $\bm{r}_{ilk}(\bm{X})$ is only one dimension, so the point-to-plane cost function of $N$ points on the $l^{th}$ plane has the following form:
{\setlength\abovedisplayskip{0.5pt}
\setlength\belowdisplayskip{0.5pt}
\begin{equation}
    \begin{split}
        \begin{aligned}
            \bm{C}^{cpp}(X) &= \frac{1}{2} \sum_{l=1}^{N} \bm{r}_{ilk}(\bm{X}) {\bm{r}_{ilk}(\bm{X})}^T 
                      = \frac{1}{2} {\bm{\pi}_l^W}^T \bm{T}_i^W \bm{G_p} {\bm{T}_i^W}^T {\bm{\pi}_l^W}
        \end{aligned}
    \end{split}
\end{equation}
}
Similar to \ref{compressed_h}, $\bm{G_p}$ is a constant $4\times 4$ matrix, which only depends on observations, not states.
Through the matrix decomposition of $\bm{G_p}=\bm{L_p}\bm{L_p}^T$, the jacobian of the new cost function can be written as 
$\bm{J}^{cpp}=\bm{L}_p^T\frac{\partial ({\bm{T}_i^W}^T {\bm{\pi}_l^W})}{\partial \bm{X}} $, and residue is $\bm{r}^{cpp} = \bm{L_p}^T{\bm{T}_i^W}^T {\bm{\pi}_l^W}$.

\vspace{-0.1cm}
\subsection{Local Plane and Point Bundle Adjustment}
When a new keyframe is inserted into the map, we perform LBA optimization. 
LBA optimizes the newest $K$ keyframe poses, IMU states, as well as the points and planes observed by these keyframes. 
Other keyframes where observe these points and planes contribute to the cost function but remain fixed in LBA.
Here, $K$ is set to 20.
We adopt the LM algorithm of Ceres\footnote{http://ceres-solver.org} to solve this minimization problem.
We set the maximum number of iterations to 10 and maximum solver time is 0.2s.
\vspace{-0.1cm}
\subsection{Global Plane and Point Bundle Adjustment}
\textbf{Loop Detection and Pose Plane Graph Optimization}
Similar to many visual SLAM~\cite{mur2017orb, campos2021orb}, we use DBoW2~\cite{galvez2012bags} to detect loop closure and use geometry information to verify their reliability.
If the loop is accepted, graph optimization is firstly used to correct large drift. 
Our optimization problem is to minimize the following energy function:
{\setlength\abovedisplayskip{0pt}
\setlength\belowdisplayskip{-2pt}
\begin{small}
\begin{equation}
    \begin{split}
        \begin{aligned}
            \bm{E}_{PPG} &= \sum_{i=1}^{N-1}\left \| ({\bm{T}_{C_i}^W}^T \bm{T}_{C_{i+1}}^W) \bm{T}_{C_{i}}^{C_{i+1}} \right \|^2 _{\bm{\Sigma}_{i(i+1)}}+         \\
                &\left \| ({\bm{T}_{C_m}^W}^T \bm{T}_{C_n}^W) \bm T_{C_m}^{C_n} \right \|^2 _{\bm{\Sigma}_{mn}} 
                 +\sum_{i=1}^{N}\sum_{l\in \mathbb{M}_i}^{}  \left \|\bm{\pi}_{l}^{W}-\bm{T}_{C_i}^{W}\bm{\pi}_{l}^{C_i}\right \|^2_{\bm{\Sigma}_{il}}
        \end{aligned}
    \end{split}
\end{equation}
\end{small}
}
Here, $N$ is the number of all keyframes. $m^{th}$ and $n^{th}$ is a pair of loop keyframes. 
$\bm{\Sigma}$ is the corresponding covariance matrix.
$\mathbb{M}_i$ is the set of planes observed by the $i^{th}$ keyframe.

\textbf{Optimization}
After pose plane graph optimization, we need to update all states including keyframe poses, IMU states, points, and planes. 
For the GBA problem, we fuse all measurements including preintegration, reprojection, compressed homography, compressed point-to-plane 
and prior plane in a tightly-coupled form. Fig.~\ref{fig:problem} shows this problem.
The optimization is to minimize the following energy function:
\begin{small}
\begin{equation}
    \begin{split}
        \begin{aligned}
            \bm{E}_{GBA} = 
            \sum_{i} \left \|\bm{r}^{IMU}_{i(i+1)}\right \|^2_{\sum_{IMU}} +
            \sum_{ik} \left \|\bm{r}^{C}_{ik}\right \|^2_{\sum_{C}} +
            \sum_{ik} \left \|\bm{r}^{C_{\lambda}}_{ik}\right \|^2_{\sum_{C_{\lambda}}} + \\
            \sum_{ijs} \left \|\bm{r}^{ch}_{ijs}\right \|^2_{\sum_{ch}} + 
            \sum_{is} \left \|\bm{r}^{cpp}_{is}\right \|^2_{\sum_{cpp}}
        \end{aligned}
    \end{split}
    \label{eq:gba}
\end{equation}
\end{small}
The first two items of Equation (\ref{eq:gba}) are similar to most VI-SLAM systems~\cite{mur2017visual,campos2021orb}.
$ \{ \bm{r}^{IMU}_{i(i+1)},~\bm{r}^{C}_{ik},~\bm{r}^{C_{\lambda}}_{ik}\}, 
\{ \bm{r}^{ch}_{ijs}\}, 
\{ \bm{r}^{cpp}_{is}  \}, 
 $ 
are respectively described in the Subsection \ref{point_measurement}, \ref{compressed_h} and \ref{compressed_pp}. 
We use LM algorithm to solve this minimization problem.
We set the maximum number of iterations to 100 and maximum solver time is 2s.
\section{EXPERIMENT}
We conduct many experiments to verify the accuracy and efficiency of our system. We perform an ablation study to test the effects of different modules and compare them with some similar algorithms. We ran our experiments on a computer with an i7-6700 CPU and 16G RAM.

\subsection{Dataset}
Our data comprises simulation data and real environment data. The simulation data is collected by AirSim~\cite{shah2018airsim}. AirSim is a simulator for drones, cars and more. We use a drone to collect data in an indoor scene of a large virtual office building. The scene is quite challenging, which is large and empty, with many textureless areas.
To verify the effect in the actual scene, we use a mobile phone on the market with ToF sensor to collect RGB, Depth, and IMU data. 
We collected the data in a small VICON\footnote{https://www.vicon.com/} room. The data in the VICON room has a high-precision trajectory provided by VICON. 
Sequences 01 to 05 are collected by AirSim, and 06 to 10 are mobile phone's data. 
The camera moves slowly on sequences 06 to 08, but rapidly on sequences 09 and 10. 

We also test with EuRoC MAV visual inertial datasets~\cite{burri2016euroc}. Our algorithm fuses with RGB-D and IMU sensor, however, EuRoC datasets have no depth images. So we generate depth images through the 3D module from~\cite{bao2022robust}. Considering the noise of the actual depth image, we generate a Gaussian random noise $n \sim {N}(0,{0.0017}^2)$. Assuming original depth is $ d $, we change it to $d^{'}=d+d\cdot n$.

\subsection{Ablation Study and Baseline}
To verify the influence of different modules of our algorithm, we consider three variants of our algorithm.
\textbf{VI-SLAM} is our basic VI-SLAM fused with RGB-D and IMU sensor, without plane-related constraints.
\textbf{VI-SLAM-P} adds planar constraints without compression on the basis of VI-SLAM.
\textbf{VI-SLAM-CP} compresses plane-related constraints on the basis of VI-SLAM-P.
\textbf{VIP-SLAM} is our complete algorithm. Different from VI-SLAM-CP, VIP-SLAM removes the states of the plane points in the optimization. 
We also compare our algorithm with ORB-SLAM2 (RGB-D)~\cite{mur2017orb},  ManhattanSLAM~\cite{yunus2021manhattanslam} and Kimera~\cite{rosinol2019kimera}. To better demonstrate the influence of the plane on the cumulative error, we also test the version without loop closure.

\begin{table}[htb]\footnotesize
    \centering
    \captionsetup{font={small}}
    \setlength{\abovecaptionskip}{0.cm}
    \setlength{\belowcaptionskip}{0cm}
    \caption{ATE RMSE(cm) of methods without/with loop closure.}
    \label{table:ATE0}
    \setlength{\tabcolsep}{0.5mm}{
    \begin{tabular}{@{}cccccccc@{}}
    \toprule
Dataset   & Seq & \begin{tabular}[c]{@{}l@{}}ORB-\\ SLAM2\cite{mur2017orb}\end{tabular}   & \begin{tabular}[c]{@{}l@{}}Manhattan-\\ SLAM\cite{yunus2021manhattanslam}\end{tabular}     
& VI-SLAM & VI-SLAM-P      & VIP-SLAM       \\ \midrule
\multirow{5}{*}{AirSim}     
& 01  & 16.8/9.1  & 12.6/X  & 5.6/2.9  & 2.2/1.7           & \textbf{1.2}/\textbf{1.4} \\
& 02  & 112.1/59.7  & 112.7/X  & 36.7/2.2  & 4.0/2.0           & \textbf{3.6}/\textbf{1.8} \\
& 03  & 15.6/9.1  & 6.5/X  & 9.3/2.7  & \textbf{2.1}/3.3  & \textbf{2.1}/\textbf{1.5}\\
& 04  & 76.6/40.1  & 420.3/X  & 156.2/16.6  & 21.1/\textbf{8.0}  & \textbf{7.8}/9.2 \\
& 05  & 63.9/62.5  & 113.4/X  & 8.4/5.1  & \textbf{1.8}/1.9  & 2.0/\textbf{1.1} \\ \midrule
\multirow{5}{*}{\begin{tabular}[c]{@{}l@{}}Mobile\\ Phone\end{tabular}} 
& 06  & \textbf{5.9}/\textbf{1.7}  & 14.8/X  & 9.3/2.3           & 9.6/2.7          & 7.9/1.8 \\
& 07  & 14.3/3.0                    & 8.3/X  & 10.0/\textbf{1.6}  & 3.4/2.7          & \textbf{3.3}/2.4 \\
& 08  & 10.4/3.6                    & 6.2/X  & 6.5/\textbf{2.3}  & 3.6/2.7          & \textbf{3.0}/\textbf{2.3}\\
& 09  & X/X                            & X/X      & 15.1/2.0           & 5.5/2.5          & \textbf{5.4}/\textbf{1.6}\\
& 10  & X/X                            & X/X      & 34.4/\textbf{2.4}  & \textbf{14.0}/2.8 & 24.5/4.0 \\ 
\midrule
& Avg & X/X                            & X/X      & 29.2/4.0           & 6.7/3.0          & \textbf{6.1}/\textbf{2.7}\\ \bottomrule
    \end{tabular}}
\end{table}

\begin{table}[htb]\footnotesize
    \centering
    \captionsetup{font={small}}
    \setlength{\belowcaptionskip}{-0.2cm}
    \caption{ATE RMSE(cm) of methods without/with loop closure.}
    \label{table:ATE1}
    \setlength{\tabcolsep}{0.5mm}{
    \begin{tabular}{@{}cccccccc@{}}
    \toprule
    Dataset & Seq & \begin{tabular}[c]{@{}l@{}}ORB-\\ SLAM2\cite{mur2017orb}\end{tabular}  
    &\begin{tabular}[c]{@{}l@{}}Manhattan-\\ SLAM\cite{yunus2021manhattanslam}\end{tabular} 
    & \begin{tabular}[c]{@{}l@{}}Kimera\cite{rosinol2019kimera}\end{tabular} & VI-SLAM      & VIP-SLAM       \\ \midrule
    \multirow{6}{*}{EuRoC} 
    & V101       & \textbf{2.2}/2.2          & 6.4/X          & 5.0/5.0   & 2.9/2.9          & \textbf{2.2}/\textbf{2.0}  \\
    & V102       & 21.0/\textbf{2.1}          & 32.9/X          & 8.0/11.0   & 11.7/5.4          & \textbf{2.2}/2.4  \\
    & V103       & 24.0/6.2                   & 7.2/X          & 7.0/12.0   & 5.2/3.1          & \textbf{2.1}/\textbf{2.2}  \\
    & V201       & 5.1/6.0                   & 7.3/X          & 8.0/7.0   & 5.4/\textbf{3.1} & \textbf{3.9}/3.2  \\
    & V202       & 5.3/4.0                   & 9.3/X          & 10.0/10.0   & 5.5/4.4          & \textbf{2.9}/\textbf{3.1}   \\      
    & V203       & X/X                           & X/X              & 21.0/19.0   & 20.5/8.1          & \textbf{7.8}/\textbf{7.5}           \\
 \midrule          
    & Avg        & X/X                           & X/X              & 9.8/10.7   & 8.5/4.5          & \textbf{3.5}/\textbf{3.4}  \\ \bottomrule
    \end{tabular}}
\end{table}

\subsection{Accuracy}
We run each sequence three times and show the average accuracy results in Tables \ref{table:ATE0} and \ref{table:ATE1}. The results of Kimera came from \cite{rosinol2019kimera}. Under the cases of weak texture and fast motion, the VI-SLAM algorithm is more accurate and robust than ORB-SLAM2 and Manhattan-SLAM, which is mainly due to the integration of IMU does not rely on external information. Both VIP-SLAM and VI-SLAM-P achieve the best accuracy, which demonstrates that compressing multiple observations and replacing the reprojection constraints of the point on the plane with homography constraints are effective. However, because of the loop closure effect, the cumulative error of SLAM is effectively eliminated, resulting in the accuracy difference between VIP-SLAM and VI-SLAM is minor. With the loop closure module disabled, VIP-SLAM and VI-SLAM-P have achieved obvious advantages on all datasets, which effectively proves that plane constraints are very helpful to ease the accumulated errors of SLAM. 

In all sequences, the performances of Manhattan-SLAM and Kimera are poor. For Manhattan-SLAM, we guess there are two reasons. 1) Our data is more challenging, with weak textures, large indoor areas, fast movements, etc. 2) Manhattan-SLAM does not make full use of plane information. Besides they only use observations from a single frame to a plane. We use the point-to-plane constraints of direct observation to achieve a more accurate estimation of single-frame observations. Moreover, we convert the reprojection of the plane point to homography constraints to establish the relationship between multiple frames and planes. As for Kimera, the poor accuracy of the plane and local plane information cause poor performance. They only consider the local planes within the small window.
\begin{table}[htb]\footnotesize
   \centering
    \captionsetup{font={small}}
    \caption{Time-consuming and corresponding accuracy of GBA for different variant algorithms}
    \label{table:cost_time_ba}
    \setlength{\tabcolsep}{0.5mm}{
    \begin{tabular}{@{}ccccc@{}}
    \toprule
                       & VI-SLAM & VI-SLAM-P      & VI-SLAM-CP    & VIP-SLAM \\ \midrule
    Iter.              & 10      & 10             & 10            & 10       \\
    Redisual (ms)      & 65      & 181            & 117           & 22       \\
    Jacobians (ms)      & 145     & 777            & 213           & 86       \\
    Linear (ms)         & 425     & 593            & 462           & 113      \\
    Pre-processor (ms)  & 24      & 84             & 27            & 3        \\
    Post-processor (ms) & 3       & 7              & 3             & 1        \\
    Sum (ms)            & 722     & 1845           & 912           & 254      \\
    RMSE (m)           & 0.038   & 0.032          & 0.027         & 0.032    \\ \bottomrule
    \end{tabular}}
\end{table}
\begin{figure}[htb]
    \centering
    \setlength{\belowcaptionskip}{-0.7cm}
    \includegraphics[width=0.8\linewidth]{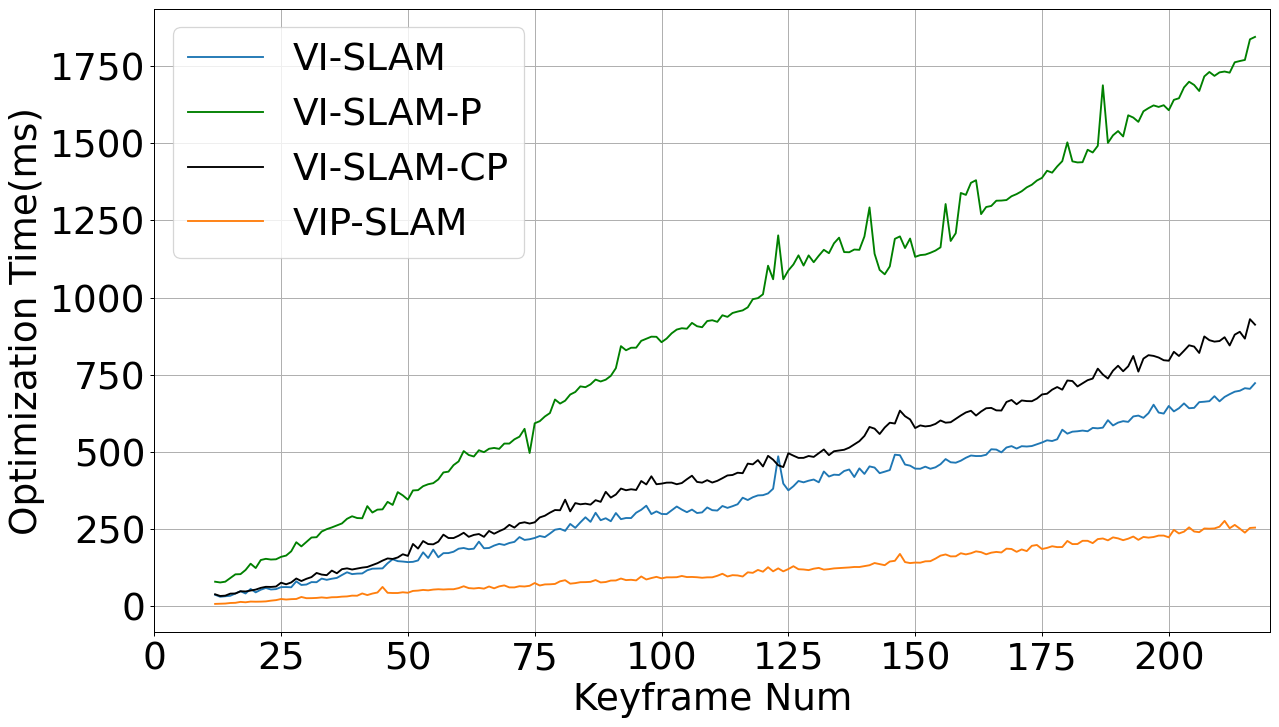}
    \captionsetup{font={footnotesize}}
    \caption{The GBA time for variants of our algorithm.}
    \label{fig:cost_time}
\end{figure}
\subsection{Runtime}
To verify VIP-SLAM cost time of optimization, we do an experiment of GBA with sequence 09. All algorithms have a maximum number of 10 iterations and do not limit the optimization time. Fig.~\ref{fig:cost_time} shows a comparison of the GBA time for variants of our algorithm. The horizontal axis is the numbers of keyframes, and the vertical axis is the optimization time of GBA. We find that the optimization time of VI-SLAM-P is much higher than VI-SLAM due to too many additional point-to-plane and homography measurements. After using the compressed measurements as VI-SLAM-CP, the optimization time is close to VI-SLAM but still higher. Finally, removing the state of plane points, optimization time is nearly 2 times faster than VI-SLAM. We also find the optimization time consumption will be reduced obviously as the number of keyframe increases. 
Table \ref{table:cost_time_ba} shows the time-consuming and accuracy of GBA for different variant algorithms. 
The states of GBA include 215 keyframes, 2,236 planar points, 2,032 non-planar points, 10 vertical planes, and 6 horizontal planes. 
We list several important sections of optimization time-consuming, as we can see, time-consuming is reduced for all important sections of our VIP-SLAM with close accuracy.
\vspace{-0.25cm}
\section{CONCLUSIONS}
Our goal is to make full use of multiple sensors' information to achieve a high-precision, lightweight and robust SLAM system. The experiments demonstrates the effectivess of the proposed system. Using plane information not only helps to reduce the cumulative error of the system but also accelerates the GBA. Our points and planes GBA is nearly 2 times faster than the sparse points based SLAM algorithm. So far, the back-end map of our system still includes a lot of points. In the future, we will explore whether a more lightweight pure plane-based visual SLAM back end is feasible.







\bibliographystyle{IEEEtran}
\bibliography{root}

\end{document}